# Parsing Thai Social Data:
# A New Challenge for Thai NLP

Sattaya Singkul
*King Mongkut's Institute of Technology Ladkrabang*
Bankok, Thailand
joeysattaya@gmail.com

Borirat Khampingyot
*Chiang Mai University*
Chiang Mai, Thailand
borirat.khampingyot@gmail.com

Nattasit Maharattamalai  Supawat Taerungruang  Tawunrat Chalothorn
*Kasikorn Labs*
Nonthaburi, Thailand
{nattasit.m} {supawat.tae} {tawunrat.c}@kbtg.tech

*Abstract*—Dependency parsing (DP) is a task that analyzes text for syntactic structure and relationship between words. DP is widely used to improve natural language processing (NLP) applications in many languages such as English. Previous works on DP are generally applicable to formally written languages. However, they do not apply to informal languages such as the ones used in social networks. Therefore, DP has to be researched and explored with such social network data. In this paper, we explore and identify a DP model that is suitable for Thai social network data. After that, we will identify the appropriate linguistic unit as an input. The result showed that, the transition based model called, improve Elkared dependency parser outperform the others at UAS of 81.42%.

## I. INTRODUCTION

Dependency parsing (DP) is a task that analyzes text for syntactic structure and relationship between words. DP could be used for improving NLP tasks such as information extraction [1], question answering [2, 3], and semantic parsing [4]. Social media are platforms that people use for communication, especially in the context of customer service support (i.e., customers reporting problems or feedback to a company's social network page). In fact, customer service support via social networks is increasingly popular among business companies. Consequently, the amount of textual data a company receives from a social network channel has also increased substantially. This also poses a challenge for the company's customer service department to analyze such massive amount of data in order to identify problems and improve their service quality. To do so, each piece of text must be extracted for customer intention as well as products or services mentioned. However, social media texts are more difficult to process than traditional texts [5] and, sometimes, they can be more difficult to understand. Moreover, there is also a challenge of syntax ambiguity because it is harder to identify sentence boundaries and grammars in Thai social language.

As shown in Table I, the first and second sentence clearly indicate that a customer wants a Chopper card. The third sentence, however, consists of two intentions from the customer: 1) he/she wants to apply for a Chopper card, and 2) he/she is looking for a Chopper card that is cuter than Rilakkuma card. Finally, the fourth sentence has the most complex structure, indicating that 1) the customer wants a Chopper card, 2) the Chopper card must come with an installment plan, 3) the Chopper card must be cuter than Rilakkuma card, and 4) he/she recalls that a Rilakkuma card can withdraw cash from cash machines. The fourth sentence consists of two intentions, three services, and two brands. Such a complex sentence requires non-trivial effort from human operators to analyze. Moreover, human operators must also aggregate all the analyzed results and generate a report regularly within limited time. Such manual process is cumbersome and yet the results might be inaccurate. Therefore, there is a need for an automated system that can interpret intentions and sentiments from such complex sentences at scale.

TABLE I.    SENTENCES IN SOCIAL DATA

| Sentence Type | Sentence |
|---|---|
| Normal | ขอบัตรช็อปเปอร์ได้มะ<br>'Can I have a Chopper card?' |
| Normal | ขอสมัครบัตรช็อปเปอร์ได้มะ<br>'Can I apply for a Chopper card?' |
| Long | ขอสมัครบัตรช็อปเปอร์ของกสิกรที่น่ารักกว่าริระคุต้มะอะ<br>'Can I apply for Kasikorn's Chopper card that is cuter than Rilakkuma?' |
| Complex | ขอสมัครบัตรกสิกรอันที่ผ่อนได้มีลายช็อปเปอร์ไหมอะที่น่ารักกว่าที่กดเงินสดที่เป็นลายริระคุมะได้ป่าว<br>'I would like to apply for Kasikorn's card that can be used to pay by installments, is there a Chopper pattern, which is lovelier than Rilakkuma that can withdraw cash?' |

In order to perform automated intention classification and sentiment analysis of complex sentences (like the ones shown in Table I), Dependency Parsing (DP) must be achieved. According to [6-8], if we do not understand relationships between words, relationships between entities in a sentence cannot be extracted. In particular, if the text consists of multiple entities (as shown in Table I: complex sentence), relationships between entities can help identify which entity should be focused (e.g., entity of "Kasikorn's card" should be focused in the complex sentence example from Table I). Therefore, the lack of Thai language DP could lead to misunderstanding in the meaning of the sentence. In fact, the lack of Thai language DP is one of the reasons why high-level Thai NLP tasks (e.g., sentiment analysis, question answering) cannot be implemented. Previous research works on DP are based on English text corpus [9-11] and hence cannot be used with Thai social network text.

Nonetheless, to solve such problem, two challenges are explored and addressed in this paper. The first challenge is to identify a suitable model for parsing Thai social data. The second challenge is to identify an appropriate linguistic unit as an input for DP. The paper addresses the first challenge by analyzing the characteristics of Thai social language. To address the second challenge, the paper proposes to use Elementary Discourse Units (EDUs) as input to conform to those linguistic characteristics. Ultimately, the experiment demonstrates interesting performance resulting from the selection of suitable models and input units.

The remainder of this paper is structured as follows; the theoretical background of the related works is reviewed in Section II. Section III describes the characteristics of Thai social data. The experiment and its results are discussed in Section IV and V, respectively. Finally, Section VI concludes the paper.

## II. Related Works

Dependency parser (DP) is a task of natural language processing (NLP) that is widely used for extracting and analyzing grammatical structure of a sentence [12, 13]. Dependency links are close to the semantic relationships needed for text interpretation [14] (e.g., dependency relation can clearly show the relationship between words.) In addition, there are two approaches normally used in the tasks of dependency parsing: transition-based and graph-based.

Transition-based DP is a process of parsing a sequence of actions (transitions) for building a dependency graph and constructing a dependency tree by scanning left-to-right (or right-to-left) through words along the sentence. There are many research works that explore this method. For example, Zhang and Nirve [15] proposed new features that achieved the Unlabeled Attachment Score (UAS) of 92.9% on Penn Treebank and 86.0% on Chinese Treebank. Those features are composed of distance, valency, unigrams, third-order and label set. Moreover, stacked LSTM is proposed for transition-based DP by Dyer et al. [16]. Their model achieved better performance in both Stanford Dependency Treebank and Penn Chinese Treebank 5.1 with the UAS of 93.1% and 87.2%, respectively. Stenetorp [17] suggested to use recursive neural networks in transition-based parsing and achieved UAS of 86.25% on CoNLL 2008 dataset.

On the other hand, graph-based dependency parsing uses a concept of node to represent each word in a sentence. A search process then starts by constructing a dependency graph to adjust the weight of each edge in the connected graph such that 1) all nodes are covered, and 2) the sum of highest scoring edges is maximized. Flanigan et al. [18] used the inspiration of graph-based parsing techniques for abstract meaning representation (AMR). Their concept achieved an F-score of 84% on the testing data of LDC2013E117 corpus [19]. Moreover, Wang and Chang [20] proposed to used Bidirectional LSTM for graph-based parsing with English Penn-YM Treebank [21], English Penn-SD Treebank [22] and Chinese Penn Treebank (CTB5) [23]. They claimed that their results achieved better performance on Penn-SD dataset (UAS of 94.08%) where the data size is four times larger than Penn-YM (UAS of 93.51%) and CTB5 (UAS of 87.55%) datasets.

Furthermore, universal dependency (UD) [24] is a framework that aims to create treebank across different human languages. Also, UD is an open community producing more than 100 treebanks in over 70 languages. UD dataset is typically used in the research works such as [25], [26], and [27]. Parallel Universal Dependencies (PUD) treebanks, which were created for the CoNLL 2017 shared task on Multilingual Parsing from Raw Text to Universal Dependencies, are multilingual treebanks taken from news domains and Wikipedia. Moreover, there is a Thai PUD that consists of 1,000 lines of sentences or 22,322 tokens of word. Because of the lacking of labeled dataset, the Thai PUD is used as one of the corpus in this work.

## III. Characteristics of Thai Social Data

It is generally known that communication channel is one of the factors that affects language usage patterns. On social media, texts have characteristics that reflect the social conversations. For this reason, the language used on social media is diverse and constantly changes according to people, topics, and situations. In this section, we discuss the key language characteristics of Thai social data that drive us to build parser for social domain.

### A. Word

Word is a linguistic unit that represents concepts [28]. In general, the concepts represent through words are meaning or grammatical functions. However, words in social domain have behaviors that are different from those in formal domain because of the rapid variation of online communication.

In terms of word form, the same word may appear in a variety of forms. A variation of word form is usually made by sound variant [29]. For example, "จัง" /caŋ1/ is changed to "จุง" /cuŋ1/, "จรุง" /cruŋ1/, or "ชรุง" /chruŋ1/.

In terms of the meaning, a number of words that appear in social domain have different meanings to the same word form that appears in formal domain. For example, in formal domain, "กาก" /kaak2/ denotes 'the rest after the good part is removed'. But in social domain, it means 'bad'. In addition, the meanings of the words that appear in the social domain are also varied according to the number of new words being added according to the behavior of language users. These words, for example "ตะมุตะมิ" /ta1.mu4.ta1.mi4/, "สายเปย์" /saaj5.pe1/, "ป๊วะ" /puaʔ4/, are all not found in the dictionary.

In terms of function, grammatical functions of some words are extended beyond those appeared in formal domain. For instance, a word "แบบ" /bɛɛp2/ 'form, model' normally functions as a subject "*แบบ*อยู่ในลิ้นชัก" 'A form is in the drawer', an object "พนักงานยื่น*แบบ*ทางอินเทอร์เน็ต" 'Employee submits forms on the internet', or a classifier "เจ้าหน้าที่เสนอทางเลือก 2 *แบบ*" 'Officer offers 2 options'. Conversely, in social domain, "แบบ" has more grammatical functions, e.g. an adverb marker "เราก็ขึ้นรถ*แบบ*งงๆ" 'I got in a car confusedly', a subordinate conjunction for adverbial clause "นางก็เดินไป*แบบ*ไม่หันกลับเลยจ้า" 'She walked without turning back', a relativizer "เขาเป็นคน*แบบ*ไม่สนโลก" 'He is a person who doesn't care about anything', a discourse marker "*แบบ*จะไปทำงานสายแล้วไง" 'being go to work late'.

With the aforementioned characteristics of the words, a language processing tool built on formal domain data may return unsatisfactory results. Because there are words that do not appear in formal domain. Moreover, the new words and the extended grammatical functions of the words that cannot be found in formal domain will directly affect the part of speech tagging task. If the function of a word changes, the POS of word changes accordingly. Especially, in a syntactic task like this work, POS plays a very important role in expressing the relationship between words in the text. For such reasons, a parser model in this work uses Thai social data for training and testing.

*B. Sentence*

Sentence structures in the social language exhibit various complexity levels. For example, each sentence may consist of a small amount of words or may contain complex clauses that modify each other. However, the complexity of the social language is different from that of the formal language. This represents a challenge for social language processing. However, the language used in online media is similar to the spoken language. In addition to the characteristics of the words mentioned in the previous section, the characteristics of sentences in the social domain are also influenced by the spoken language. For this reason, the sentence structure is not strict. Consequently, many sentences cannot be communicated clearly.

For example, a sentence "แม่ชอบไปพารากอนมีหลายชั้น" 'Mom likes to go to Paragon has many floors'. This sentence informs two ideas: "Mom likes to go to Paragon" and "Paragon has many floors". As usual, in formal style, this sample sentence should be written separately into 2 sentences. It is not known how this phenomenon occurs, but this could be assumed that ellipses are the mechanism behind them. The ellipsis is a linguistic mechanism that is often used in spoken language [30], which includes languages in social domain.

Considering the example sentence above, there is possible that a relativize "ซึ่ง" was removed form "แม่ชอบไปพารากอน (ซึ่ง) มีหลายชั้น" 'Mom likes to go to Paragon (which) has many floors'. Based on this assumption, the other types of grammatical units that can be removed in social languages are found. For example, a verb "ผม(คิด)ว่าเขาไม่ไปหรอก" 'I (think) that he doesn't go', or a complementizer "แบงก์ชาติ คาด(ว่า)อัตราเงินเฟ้อของไทยมีแนวโน้มต่ำลง" 'Bank of Thailand expects (that) Thailand's inflation rate will be lower'

The complexity of the sentence is another characteristic that needs to be discussed. Since sentences in social media are not formal and are similar to spoken language. The length of sentences is vary. Most of sentences are very complex because the speaker typed the sentence immediately without proper screening for clear communication. Therefore, they may be complex and difficult to understand. For example, a sentence "ช่วยโพสรูปที่ถ่ายเมื่อวานตอนเย็นที่เราไปกินข้าวกันที่ สยามสแควร์ที่เรานั่งข้างๆ เธอให้ทีได้ม้ะ" 'Can you post a photo taken yesterday evening that we went to have dinner together at Siam Square that I sat next to you?'. This sentence consists of at least 4 main information: "Can you post a photo?", "A photo taken yesterday evening", "we went to have dinner together at Siam Square on yesterday evening", and "a photo that I sat next to you".

However, because of the complex modification of this sentence, some people may receive more information, i.e. "yesterday evening that I sat next to you". Such ambiguity is common in social language, and it has a significant effect on processing. Actually, a clause "ที่เรานั่งข้างๆ เธอ" 'that I sat next to you' can modify many nouns in the sentence, including "รูป" 'picture', "เมื่อวาน" 'yesterday', "ตอนเย็น" 'evening', "เมื่อวานตอนเย็น" 'yesterday evening', and "สยามส แควร์" 'Siam Square'. Therefore, the information received will depend on the noun that the hearer selects to modify.

With problems of Thai sentence mentioned above, together with the characteristics of the Thai language in which the sentence has no clear boundary [31], EDUs [32] is chosen to be a processing unit in this work. Due to, in semantic perspective, EDUs can convey single piece of information clearly. On the other hand, in syntactic perspective, EDUs are in the form of clauses or phrases with a strong marker [33] and hence can be clearly identified the boundaries. Furthermore, because of a characteristic of clauseness, structure of EDUs is also less complicated than sentences.

IV. EXPERIMENT

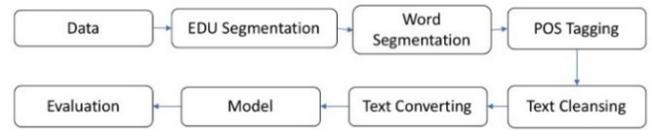

Fig. 1. Experiment process

*A. Data*

There are 2 Thai datasets used in the experiment: public UD data and social data in financial domain. Both datasets contain 1,000 sentences are grouped into 10 folds for cross-validation. Each fold consists of 800 sentences, 100 sentences and 100 sentences, respectively.

*1) Thai Social data*

This dataset is collected from social media, such as Facebook, Twitter, and Pantip by focusing on financial domain. The data is analysed and segmented into EDU by applying the principles proposed by Intasaw and Aroonmanakun [33]. The size of dataset is 219,585 EDUs.

In term of the length of EDUs, as shown in Fig. 2, the distribution of word per EDU varies. The length of word per EDU is between 2-24 words and has uniform distribution.

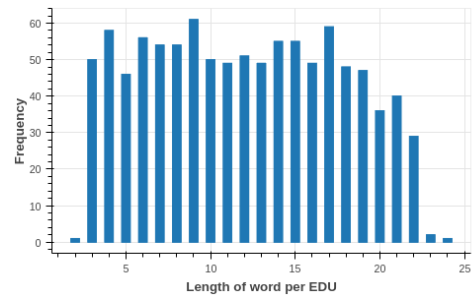

Fig. 2. distribution of word length

Fig. 3 shows the distribution of POS and the number of POS tag sets used in the data. The tag set is adapted from a universal POS tag set [34].

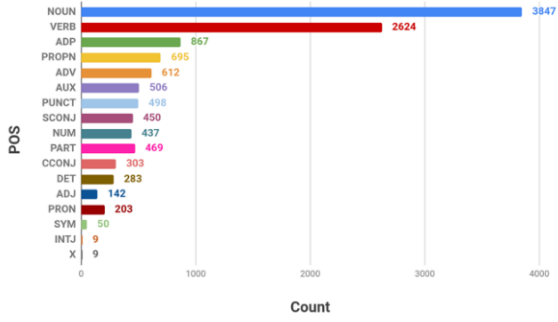

Fig. 3. Distribution of POS

*2) UD Thai Tree Bank*

This dataset consists of 22,322 words, which is a standard Thai language dataset normally used in supervised learning. The label of each sample is the relationship between words.

### B. Evaluation

There are many different evaluation metrics used in the dependency parsing task. The commonly used metrics are unlabeled attachment score (UAS) and labeled attachment score (LAS). However, due to the lack of dependency labels, UAS is used to evaluate the quality of dependency parsers. UAS focuses on the percentage of words that get the correct predictions. Equation (1) defines UAS as the number of correctly predicted heads divided by all heads in ground truth.

$$UAS = \frac{\text{\# of correct heads}}{\text{\# of heads}} \quad (1)$$

### C. Preprocessing

Before the training step, the dataset is passed through data preprocessing step and turned into an appropriate format. There are 5 processes : EDU segmentation, word segmentation, part of speech tagging, text cleansing, and text converting. The sentences are segmented in EDU segmentation process and then word segmentation process. After that, each word is marked into a category of words such as subject, verb, noun and objective. Next, special characters and excessive space are removed. Then, the text and number that have been separated by the error of word segmentation are combined in the text cleansing process. Furthermore, text converting process will convert the data to a universal format (CONLL-U format).

### D. Model

Transition-based and graph-based methods explored and used in the experiment are discussed in this section.

*1) Transition-based*

Transition-based models identify relationship between words by considering the transition of words through oracle parsing in order to see the change in each transition as shift reduction. They then use mapping features for extracting feature and converting the data into a suitable format for model training.

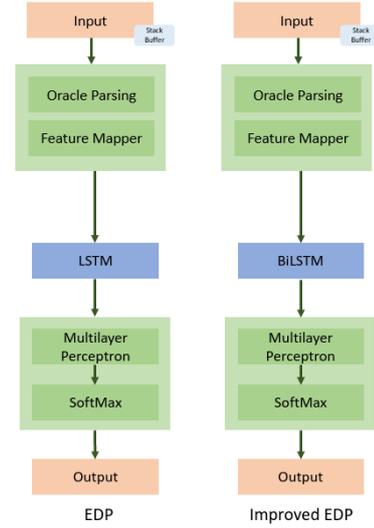

Fig. 4. A structure of transition based models consist of Elkaref Dependency Parser (EDP) and Improved Elkaref Dependency Parser.

*a) Elkaref Dependency Parser (EDP)*

Elkaref Dependency Parser [35] is composed of a single LSTM hidden layer replacing the hidden layer in the usual feed-forward network architecture. It also proposes a new initialization method that uses the pre-trained weights from a feed-forward neural network to initialize the LSTM-based model.

*b) Improved Elkaref Dependency Parser*

The concept of EDP, which has only one direction of word sequence relation, may not be enough. The concept of Kiperwasser Dependency Parsing [36] is developed using Bi-LSTM instead of LSTM to extract bi-directional features word sequence relation. Each sentence token is associated with a Bi-LSTM vector representing the token in its sentential context. Feature vectors are then constructed by concatenating a few Bi-LSTM vectors.

*2) Graph-based*

Graph-based models identify relationship between words by considering the characteristics of the graph. They focus on each pair of words and check if the pair correlate through the matrix scoring process using LSTM Encoder and Decoder.

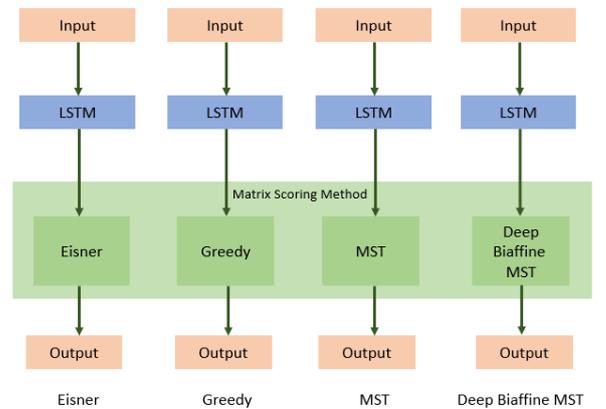

Fig. 5. A structure of graph-based models consists of Eisner, Greedy, Maximum Spanning Tree (MST) and Deep Biaffine MST.

*a) Eisner*

Eisner is a bottom-up dependency parsing algorithm. It is a projective dependency parsing and focuses on subgraph process. Adding one link at a time making it easy to multiply the model's probability factor similar to CKY method.

*b) Greedy*

Greedy is an algorithm which always selects the highest weighted edges. It is non-projective dependency parsing and compares on next-to-edge by memory-based parser [37].

*c) Maximum Spanning Tree (MST)*

Maximum Spanning Tree finds a dependency tree with higher score on a directed graph. Scores are independent from other dependencies. It is a non-projective dependency parsing and applied Chu-Liu-Edmonds algorithm [37] to find MST from directed graphs. There are composed of 3 steps: greedy, contract and recursive. Greedy step finds edges with the highest weight. Contract step detects cycles and breaks them by removing the edge with the smallest value in the cycle. Recursive step repeats the process until a spanning tree is obtained.

*d) Deep Biaffine MST*

Deep Biaffine MST [38] is a deep learning model. By adding the Bi-LSTM and a Biaffine classifier, the model performs comparably to the state-of-the-art model. The model utilizes Bi-LSTM, which gives a long-term dependency, and Biaffine classifier, which improves parsing speed.

V. RESULTS AND DISCUSSIONS

There are two experiments in this paper. The first experiment focuses on the correlation between word length and error rate. The second experiment focuses on Thai social language model.

*1) The correlation between word lengths and error rates*

This experiment was conducted on both UD dataset and Social Banking domain dataset. Two different types of methods, transition-based and graph based, are used to analyze how word lengths affect the model performance. The Improved Elkaref Dependency Parser [35] is used for training a transition-based model and the Deep Biaffine Attention [38] is used for training a graph-based model. The mean error rate is evaluated by counting the frequency of the wrong predictions in each sentences / EDUs and calculating mean error rate in each word length. Fig. 6 (a) and (c), representing training and testing on UD dataset with transition-based and graph-based, show that the more number of words in the sentences or EDUs are contained, the more error rates are found for both transition-based and graph-based model. In addition, Fig. 6 (b) and (d), which is training and testing on Social Banking domain dataset, show that nine out of ten folds yield the same direction of correlation between word lengths and error rates. Due to Social Banking domain dataset separated into EDUs with short words, this might cause a different correlation result in another fold. To simplify the problem, EDU segmentation is suggested to be used in Thai dependency parsing instead of sentence segmentation.

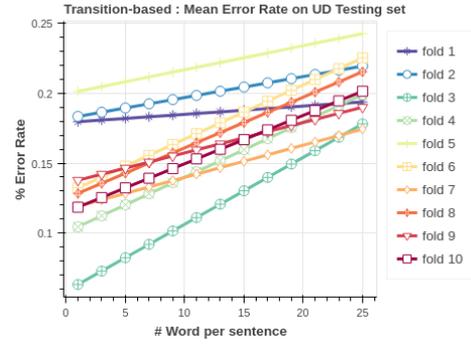

(a)

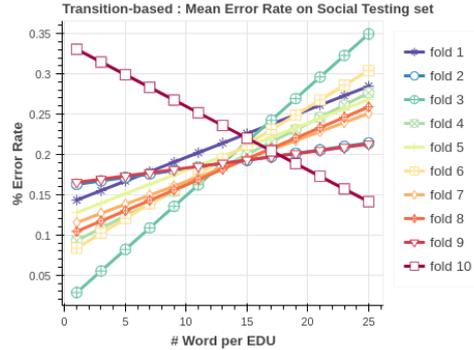

(b)

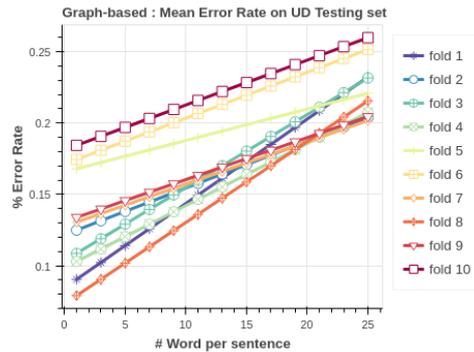

(c)

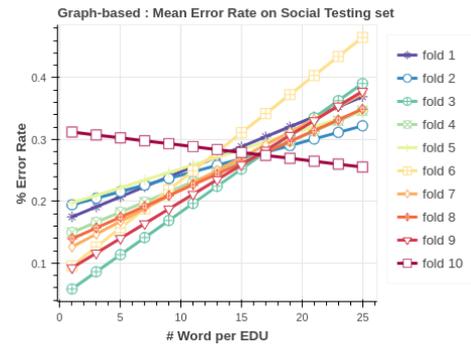

(d)

Fig. 6. Each line represents a direction of error rate occurred while the number of word per EDU increase using linear regression in (a) UD Testing set with transition-based, (b) Social Testing set with transition-based, (c) UD Testing set with graph-based and (d) Social Testing set with graph-based. X axis represents the number of words. Y axis represents percent error rates. Transition-based and graph-based use the same data distribution of UD and social dataset in training, validation and testing.

## 2) Thai Social Model

This experiment was conducted to find the best model for Thai social dependency parsing. As shown in Table 2, there are six models, including two transition-based models and four graph-based models, tested in this experiment. The results show that transition-based models perform better than graph-based models. The improved Elkaref dependency parser achieves the average 10-fold UAS of 78.62% on UD dataset and the average 10-fold UAS of 79.84% on social dataset. Moreover, the transition-based UAS (79.84%) outperformed the graph-based UAS (73.27%) on social dataset. Because of the concept of EDUs, the word length in sentence is always longer than or equal to the word length in EDU. This finding is consistent with the research work in [39], which found that "transition-based models performed better than graph-based models at short length sentences". The best model for Thai social model is improved Elkaref dependency parser.

TABLE II. RESULT ON THE UD DATASET AND SOCIAL DATASET

| Type | Model | UD Dataset | Social Dataset |
|---|---|---|---|
| | | UAS | UAS |
| Transition | EDP | 55.01 | 73.92 |
| | Improved EDP | **78.62** | **79.84** |
| Graph | Eisner | 56.93 | 58.37 |
| | Greedy | 55.24 | 53.80 |
| | MST | 57.12 | 60.99 |
| | Deep Biaffine MST | 76.95 | 73.27 |

## VI. CONCLUSION

In this paper, we have shown that length is one of the error factors in the dependency parsing problem. We suggested the use of EDU segmentation to simplify sentences instead of using sentence segmentation or long raw text. Our experimental results also show that transition-based DP models outperform the graph-based DP models in Thai social data when segmented by EDUs. Moreover, improved Elkaref dependency parser yielded the best performance among various DP models. For future works, exploration of error factors is a promising area to explore in order to improve the model performance.

## VII. ACKNOWLEDGEMENT

This work was supported by Kasikorn Business-Technology Group (KBTG).